\def\year{2022}\relax
\documentclass[letterpaper]{article} 
\usepackage{aaai22}  
\usepackage{times}  
\usepackage{helvet}  
\usepackage{courier}  
\usepackage[hyphens]{url}  
\usepackage{graphicx} 
\usepackage{natbib}  
\usepackage{caption} 
\DeclareCaptionStyle{ruled}{labelfont=normalfont,labelsep=colon,strut=off} 
\usepackage{algorithm}
\usepackage{algorithmic}

\usepackage{newfloat}
\usepackage{listings}
\usepackage{multirow}
\floatstyle{ruled}
\newfloat{listing}{tb}{lst}{}
\floatname{listing}{Listing}
\usepackage{amsmath}
\usepackage{array}
\newcolumntype{P}[1]{>{\centering\arraybackslash}p{#1}}

\title{UNITER-Based Situated Coreference Resolution with Rich Multimodal Input}
\author {
    Yichen Huang,\textsuperscript{\rm 1}
    Yuchen Wang, \textsuperscript{\rm 1}
    Yik-Cheung Tam \textsuperscript{\rm 1}
}
\affiliations {
    \textsuperscript{\rm 1} New York University Shanghai\\
    yh2689@nyu.edu, yw3642@nyu.edu, yt2267@nyu.edu
}

\usepackage{bibentry}

\begin{document}

\maketitle

\begin{abstract}
We present our work on the multimodal coreference resolution task of the Situated and Interactive Multimodal Conversation 2.0 (SIMMC 2.0) dataset as a part of the tenth Dialog System Technology Challenge (DSTC10). We propose a UNITER-based model utilizing rich multimodal context such as
textual dialog history, object knowledge base and visual dialog scenes
to determine whether each object in the current scene is mentioned in the current dialog turn. 
Results show that the proposed approach outperforms the official DSTC10 baseline substantially, with the object F1 score boosted from 36.6\% to 77.3\% on the development set, demonstrating the effectiveness of the proposed object representations from rich multimodal input.
Our model ranks second in the official evaluation on the object coreference resolution task with an F1 score of 73.3\% after model ensembling.
\end{abstract}

\section{Introduction}
The goal of Situated and Interactive Multimodal Conversation (SIMMC) 2.0~\cite{simmc2} is to aid the conversational AI community in developing successful multimodal assistant agents capable of handling real-world multimodal dialog inputs. As part of the DSTC10 challenge, this dataset includes dialogs and multimodal contexts in the fashion and furniture domain that closely resemble real-world scenarios with more complex and cluttered images compared to previous datasets~\cite{simmc, simmc_platform}.

In this paper, we focus on the multimodal coreference resolution sub-task. Given a multimodal dialog context including dialog history, raw scene images, bounding boxes of detected objects and their coordinates, scene graphs, and a knowledge base (KB) of objects, our task is to determine whether each object is mentioned in the current user dialog turn. For example, a system turn asks, "which group of pants are you referring to?" and a user may reply 'The ones on the left.', requiring to resolve which object IDs are referred to from a given active scene image.
This task necessitates the system's understanding from contextualized multiple modalities, including textual, visual and structural object KB, laying the foundation for downstream processing, including dialog state tracking, response generation and retrieval.

Top approaches in the previous SIMMC challenge \cite{simmc_sol1, simmc_sol2} cast action prediction, dialog state tracking and response generation as a sequence-to-sequence generative approach where the multimodal context is flattened as a sequence of tokens which are coupled with dialog history for encoding. An auto-regressive decoder generates the corresponding actions, dialog state and response as a sequence of tokens. While such approaches were effective in the last challenge, they are incompatible with our task for several reasons. Firstly, it is not trivial to flatten raw images and scene graphs as a sequence of input symbols. Secondly, the flattened multimodal context significantly increases the input sequence length in SIMMC 2.0, where a scene image can contain more than 20 objects, easily exceeding the typical input limit of 512 tokens. Thirdly, the generated output sequence relies on ad-hoc post-processing to make sure that, for example, the object IDs are in the correct order and without duplicates during the beam search decoding. 

We propose a UNITER\cite{uniter}-based model for SIMMC 2.0. UNITER is proposed in computer vision (CV) for universal embeddings for image and text. To achieve this goal, UNITER is pre-trained with masked language modelling, masked region modelling and word-region alignment criteria on parallel image-text data. We extend UNITER to handle complex multimodal inputs. For object coreference resolution, we focus on the rich feature representation of each object to enable the underlying transformer model to comprehend the coreferences between textual dialog history with object candidates. In particular, each object is modelled with object index embedding, image embedding from a deep pre-trained CV model, KB entries cast as prompts for sentence embedding and additional feature engineering such as whether an object was mentioned in previous system dialog turns.
Motivated from prior work of using scene graphs for visual question answering~\cite{DBLP:journals/corr/abs-2101-05479}, we also
incorporate scene graphs that include the positional relationship between objects in a scene. In particular, we evaluate two methods: 1) injecting scene graph information as attention biases; 2) through additional relation-aware self-attention layers. Finally, each object candidate is treated as an input embedding into UNITER and outputs a binary object mention label. In other words, our model contains a binary classification head per object candidate. 


The paper is organized as follows: Section~\ref{sec:simmc2} provides an overview of the multimodal coreference task and the SIMMC 2.0 dataset. Section~\ref{sec:proposed} presents our proposed approaches. Section~\ref{sec:expt} describes experiments and results, with concluding remarks in section~\ref{sec:conclude}.

\section{Task Descriptions}
\label{sec:simmc2}
The SIMMC 2.0 dataset assumes scenarios in shopping settings where a user and an assistant agent co-observe scenes. The dataset is collected through a VR scene generator and is highly structured with data on different levels (turn, scene, dialog and KB metadata). A dialog can involve multiple scenes and turns. On the dialog level, the dataset contains the bounding box ids of all objects mentioned in the entire dialog. On the turn level, the dataset contains utterances and dialog state annotations (including the bounding box ids of objects mentioned in the current utterance) of both the user and the system and the scene id. On the scene level, the dataset contains the raw image, each object's bounding box, the relationships between objects (e.g. left, right) and unique ids linking each object to a knowledge base. Each knowledge base entry contains non-visual metadata (e.g. brand, price, available sizes) and visual metadata (e.g. type, color, pattern, sleeve length). The dataset contains a total of 11.2k dialogs and 1.5k scenes. The dataset is split into four sets: train (65\%), dev (10\%), dev-test (10\%) and test-std (15\%). Test-std is a hold-out hidden set for the DSTC10 challenge. The dataset involves two domains: fashion (7.2k dialogs) and furniture (4k dialogs).

The task of multimodal coreference resolution is to resolve scene-level ids in user utterances. An utterance can involve multiple references, in which case the ground-truth output is an unordered list of bounding box ids. The allowed inputs at inference time include past utterances of a user and the system, the current-turn utterance of the user, past object mentions of the system, scene data and non-visual metadata. Note that all user object mentions and all dialog state annotations are not allowed at inference time. Utterances with ambiguities are not included. The performance is evaluated using object F1.

\section{Proposed approach}
\label{sec:proposed}
\begin{figure*}[t]
\centering
\includegraphics[width=1.9\columnwidth]{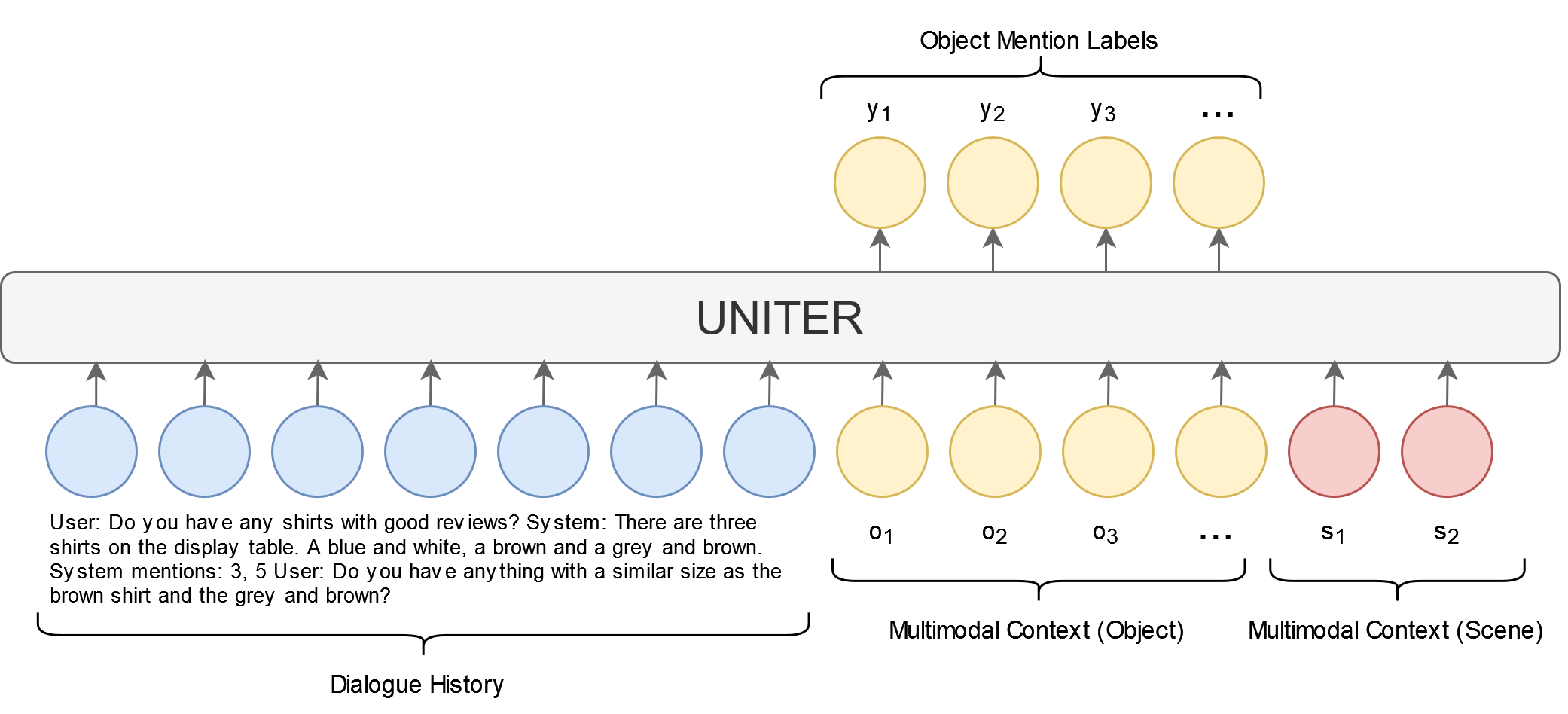} 
\caption{An overview of the proposed model.}
\label{model_overview}
\end{figure*}

We formulate multimodal coreference resolution as an instance of binary token classification. Given dialog history $U$, object embeddings $O=o_1o_2...o_I$ and scene embedding $S=s_1...s_j...s_J$, we aim to predict binary object mention labels $Y=y_1y_2...y_I$ indicating whether each object $o_i$ is mentioned in the current user utterance. We use a UNITER encoder \cite{uniter} to encode the above inputs. An overview of our proposed model is shown in Figure \ref{model_overview}.

\begin{figure}[t]
\centering
\includegraphics[width=0.9\columnwidth]{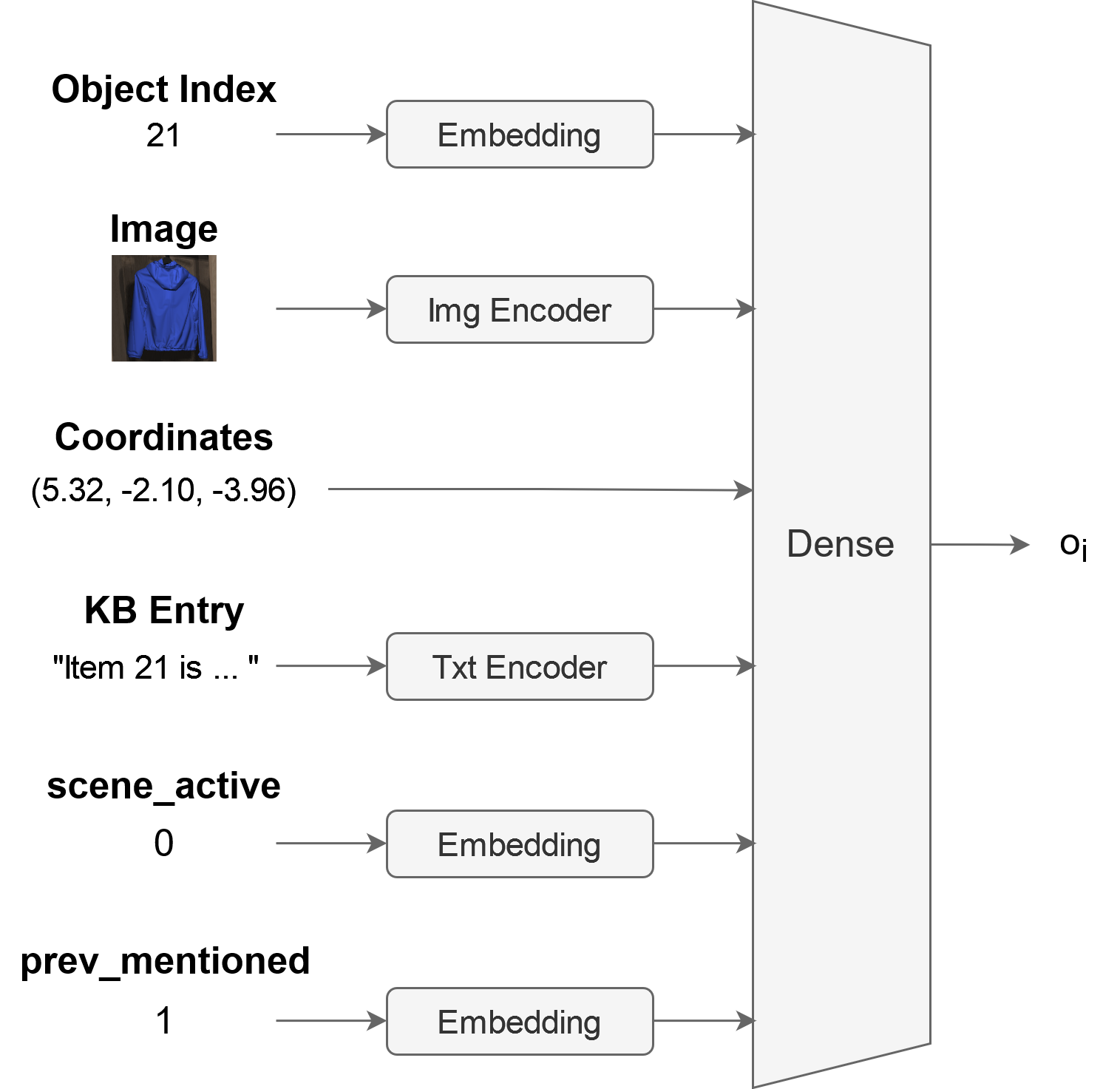} 
\caption{The separate embeddings of multimodal object features are concatenated and aggregated through a dense layer.}.
\label{obj_embedding}
\end{figure}

\subsection{ObjectEmbeddings}
As shown in figure \ref{obj_embedding}, we first obtain separate embeddings for each of the multimodal object features and then aggregate them using a dense layer. We process the features as the following:
\begin{itemize}
    \item 
    A scene-level object index is embedded through an embedding layer.
    \item
    A cropped object image is fed into a visual encoder to extract the pooled region of interest (ROI) feature. In our experiments, we use pre-trained CLIP \cite{clip} and BUTD \cite{butd} based on Faster R-CNN \cite{faster_rcnn}.
    \item
    The x, y and z coordinates of an object are used with no processing.
    \item
    The {\em non-visual} knowledge base (KB) entry of an object is first transformed into natural language form using a template shown in Table \ref{tab:templates}. Then the descriptive sentence is encoded with a text encoder such as BERT \cite{bert} and sentence BERT (SBERT) \cite{sbert}.
    \item
    scene\_active is a binary feature indicating whether an object is in the currently active scene. The label is embedded through an embedding layer. 
    \item
    prev\_mentioned is a binary feature indicating whether the system has mentioned an object in previous dialog turns. The label is embedded through an embedding layer.
\end{itemize}
These features are then concatenated and passed into a dense layer. Except for the image and text encoders, the rest embeddings are trained in an end-to-end manner.

\begin{table}
\begin{center}
\begin{tabular}{ c p{60mm} } 
 \hline
 Fashion & Item 15 is located at x : 5.32, y : -2.10, z: -3.96. It is located in the bounding box 104 334 260 133. Its price is \$59.99. Its size is XL. Its brand is Downtown Stylists. It has a customer review of 4.1 out of 5. It is available in sizes S and XL. \\ 
 \hline
 Furniture & Item 7 is located at x : -756.50, y : 0.00, z: -358.20. It is located in the bounding box 838 383 45 30. Its price is \$549. Its brand is River Chateau. It is made with metal. It has a customer review of 4.2 out of 5. \\ 
 \hline
\end{tabular}
\caption{\label{tab:templates}Examples of knowledge base entry templates for the fashion and furniture domains.}
\end{center}
\end{table}

\subsection{Scene Embeddings}

We add scene embeddings to reflect visual information not included in the object bounding boxes (e.g. relative positions and scene layouts). Scene embeddings $S=s_1...s_j...s_J$ are obtained similar to the object embeddings except we use scene-level input indicators for index, scene\_active and prev\_mentioned different from object embeddings. Otherwise, we feed the entire scene image for image encoding. Coordinates and KB encodings are not used in scene embeddings.

\subsection{Multimodal Encoder}
We use a pre-trained UNITER model to encode dialog history, object embeddings and scene embeddings. The hidden states corresponding to each object position is passed into a dense layer to produce the output logits $Z$ followed by Sigmoid function $\sigma(Z)$ for binary classification:
\begin{equation}
\begin{aligned}
    H & = \text{Encoder}(U,O,S)\\
    Z & = \text{Dense}(H)\\
    \hat{Y} & = \sigma(Z)
\end{aligned}
\end{equation}

\subsection{Scene Graph Integration}
Scene graph is a way to represent the relationship of objects in an image using a directed graph where each node in a graph represents an object and a directed edge represents a relationship. The use of scene graphs has lead to improved performance in various vision-language tasks including visual QA \cite{DBLP:journals/corr/abs-2101-05479}, image captioning \cite{AE_sg_imgcap, sg_imgcap}, text-to-image generation \cite{img_gen} and referring expression comprehension \cite{ref_exp_comp}.
This motivates us to adopt scene graphs for multimodal coreference resolution. We explore two approaches to modifying the UNITER's transformer layers via a self-attention mechanism between object pairs to incorporate scene graphs.

\subsubsection{Attention Bias}
\begin{figure}[t]
\centering
\includegraphics[width=0.8\columnwidth]{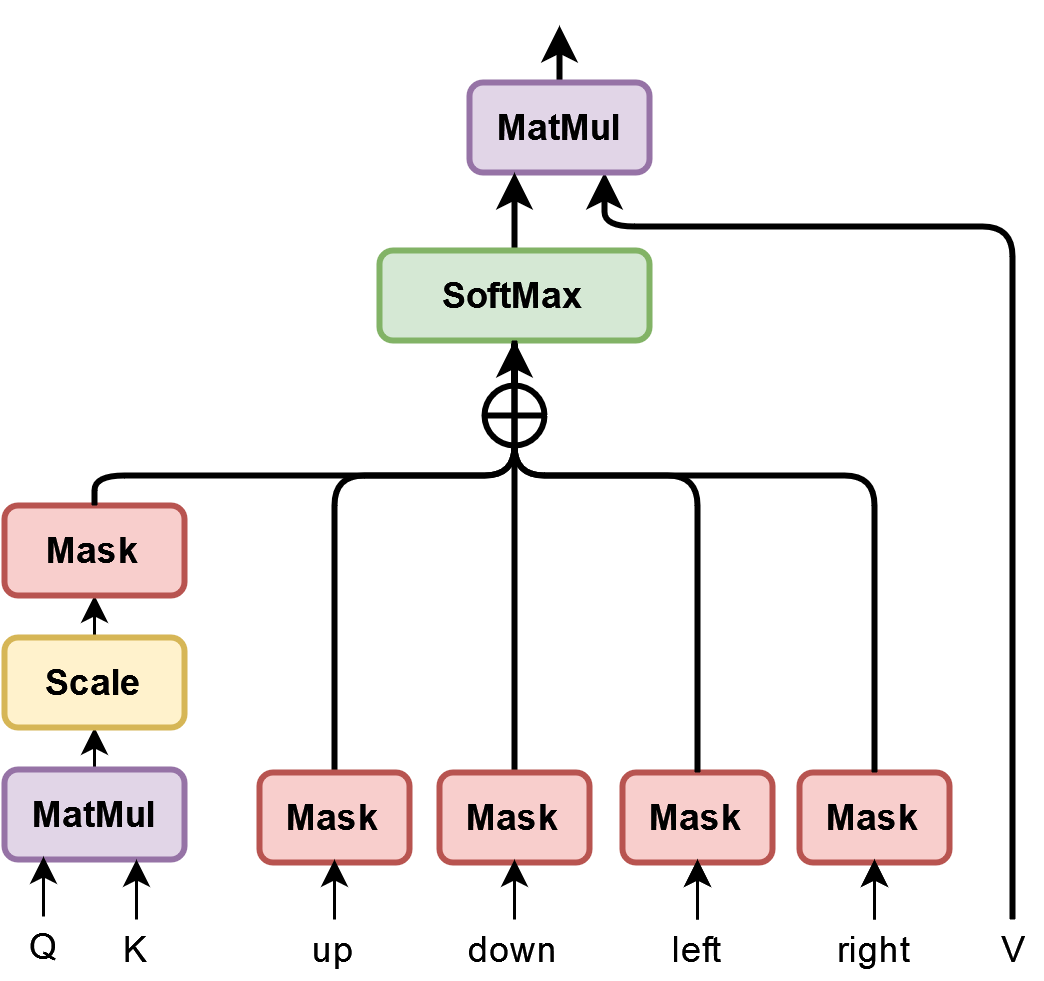} 
\caption{Integrating object-object relationship via the attention bias.}
\label{attn_bias}
\end{figure}

Similar to \cite{lambert}, we introduce a bias term modifying the attention score. For each attention head in each self-attention layer, we train a scalar embedding $\beta^r$ for each of the relationships in the scene graphs (left, right, up and down). The modified attention head is as follows (See Figure \ref{attn_bias}):
\begin{equation}
\begin{aligned}
    \alpha &= \frac{(h^{l-1}W^Q)(h^{l-1}W^k)^\top}{\sqrt{d_k}}\\
    \alpha' &= \alpha + \sum_{r=1}^n \beta^r g^r\\
    h^{l} & = \text{softmax}(\alpha') (h^{l-1}W^V)
\end{aligned}
\end{equation}
where $W^K$, $W^Q$ and $W^V$ are model parameters, $d_k$ is the dimension of the query, key and value vectors, $h^{l-1}$ and $h^l$ are the hidden states of the previous and current transformer layers respectively, $n=4$ in our case and $g^r$ is a binary mask between objects $o_i$ and $o_j$ for each of the four relationships $r$=\{up,down,left,right\}:
\begin{equation}
\begin{aligned}
    g_{ij}^r = 
    \begin{cases}
      1 & \text{if objects at position $i$ and $j$} \\
        & \text{satisfies the relationship $r$}\\
      0 & \text{otherwise}
    \end{cases}       
\end{aligned}
\end{equation}

\subsubsection{Relation-aware Self-Attention}
\begin{figure}[h!]
\centering
\includegraphics[width=0.45\columnwidth]{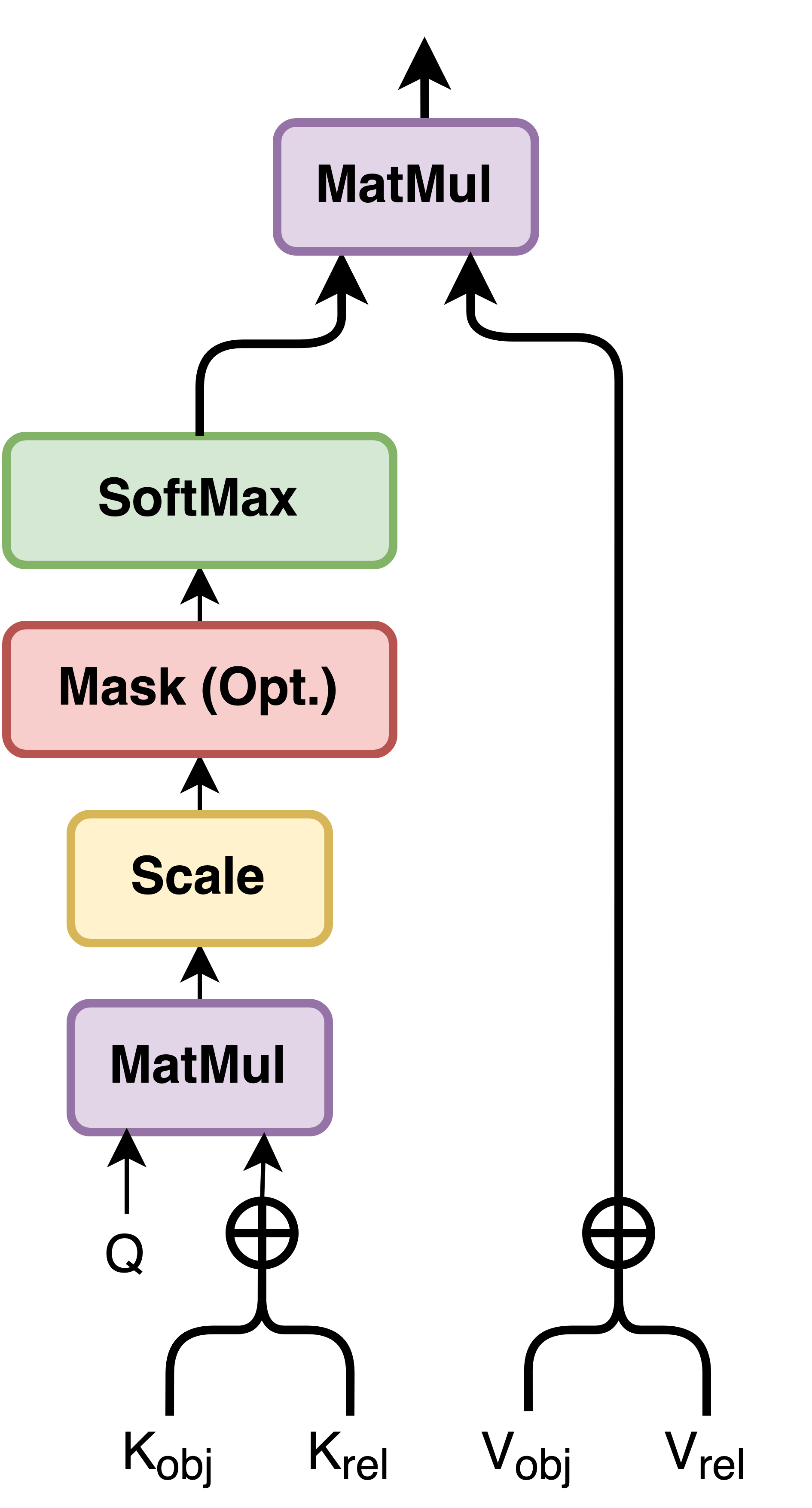} 
\caption{Relation-aware self-attention mechanism.}
\label{rel_awr_self_attn}
\end{figure}

We also experiment with relation-aware self-attention \cite{rel_aware_self_attn} where the attention is modified as follows (See Figure \ref{rel_awr_self_attn}):
\begin{equation}
\begin{aligned}
    u^r &= \text{softmax}(\frac{(h^{l-1}W^{QS})(h^{l-1}W^{KS} + g^r_{ij}W^{KR})^\top}{\sqrt{d_k}})\\
    h^{l} & = \sum_{r=1}^n g^r \circ u^r (h^{l-1}W^{VS} + g^r_{ij}W^{VR})
\end{aligned}
\end{equation}
where $W^{KS}$, $W^{KR}$, $W^{QS}$, $W^{VS}$ and $W^{VR}$  are model parameters. $\circ$ denotes element-wise multiplication so that only hidden values corresponding to objects involved in a relationship is updated. Following \cite{jointgt}, we add a relation-aware self-attention layer after every vanilla self-attention layer and add a residual connection to combine the outputs of the two.

\section{Experiments}
\label{sec:expt}
We trained the proposed model using the focal loss \cite{focalloss} with $\gamma = 2$ and $\alpha = 1$ for the negative class (i.e. an object is not mentioned), and $\alpha = 5$ for the positive class (i.e. an object is mentioned). We used the Adam optimizer \cite{adam} with a learning rate of $5 \times 10^{-6}$ and $\epsilon = 10^{-8}$. We used a batch size of 16. We trained the model for a maximum of 30 epochs and performed early stopping according to the F1 score on the official development set. Our source code can be found at \url{https://github.com/i-need-sleep/MMCoref_Cleaned}.

\begin{table}
\begin{center}
\begin{tabular}{  p{60mm} | c  } 
 \hline
 {\bf Model} & {\bf Object F1} \\
 \hline
 \hline
 GPT-2 Baseline & 0.366 \\
 \hline
 \hline
 UNITER + Faster RCNN & 0.557 \\
 \hline
 UNITER + CLIP & 0.524 \\
 \hline
 UNITER + both (Faster RCNN+CLIP) & 0.563 \\
 \hline
 UNITER + both + object idx & 0.551 \\
 \hline
 UNITER + both + object idx + coordinates & 0.579 \\
 \hline
 UNITER + both + object idx + coordinates + active\_scene & 0.582 \\
 \hline
 UNITER + both + object idx + coordinates + active\_scene + KB (BERT) & 0.665 \\
 \hline
 UNITER + both + object idx + coordinates + active\_scene + KB (SBERT) & 0.621 \\
 \hline
 UNITER + both + object idx + coordinates + active\_scene + KB (BERT+SBERT) & {\bf 0.674} \\
 \hline
 LXMERT + Faster RCNN & 0.585 \\
 \hline
 LXMERT + CLIP & 0.590 \\
 \hline \hline
 Ensemble (Configuration for submission) & {\bf 0.741} \\
 \hline \hline
 {\bf Post-eval improvement} & \\
 \hline
  UNITER* & 0.674 \\
  \hline
 UNITER* + prev\_mention & 0.728 \\
 \hline
 UNITER* + prev\_mention + attn bias & 0.734 \\
 \hline
 UNITER* + prev\_mention + relation-aware self-attn & 0.733 \\
 \hline
 LXMERT* + prev\_mention & 0.658\\
 \hline \hline
 Ensemble & {\bf 0.773}\\
 \hline
\end{tabular}
\caption{\label{tab:eval}Multimodal coreference resolution performance on the dev-test split. For models with both image or text encoders, the encodings are concatenated together. The ensembled output for submission is based on the top five UNITER-based models. The ensembled output post-evaluation is based on the best LXMERT-based model and the top three UNITER-based models. UNITER* and LXMERT* denote the best configuration used during DSTC10 evaluation.}
\end{center}
\end{table}

\begin{table}
\begin{center}
\begin{tabular}{  P{40mm} | P{10mm}  } 
 \hline
 {\bf Entry} & {\bf Object F1} \\
 \hline
 \hline
 Team 4 & 0.758 \\
 \hline
 {\bf Team 9 (Ours)} & 0.733 \\
 \hline
 Team 8 & 0.682 \\
 \hline
 Team 10 & 0.682 \\
 \hline
\end{tabular}
\caption{\label{tab:std}Official results for multimodal coreference resolution on the held-out test-std split.}
\end{center}
\end{table}

Table \ref{tab:eval} summarizes the evaluation result of the baseline models and our models with different inputs. The organizer provided the GPT-2 baseline, treating the input and outputs as a flattened sequence of tokens. 
We also compared the results with the same approach by replacing UNITER with LXMERT \cite{lxmert}. Our UNITER-based models outperformed the baseline by a large margin by simply using a basic UNITER model with Faster RCNN or CLIP for image embeddings. As we incrementally incorporated additional features such as object coordinates, "is an object inside the active scene?" (active\_scene), KB entity description via BERT and SBERT, the object F1 score was further improved from 0.557 to 0.674 on the official devtest split, which was the best single model before the official evaluation deadline. Notably, UNITER-based models outperformed LXMERT-based ones under the same configuration. We ensembled the top five UNITER-based models for our DSTC10 submission. Due to the time constraint, some of the models used in the ensemble were not fully trained. 

Table \ref{tab:std} shows our entry in the DSTC10 challenge on the test-std set among the top teams—we ensemble five models with different input settings. Finally, we achieved 2nd place in the evaluation on multimodal coreference resolution subtask. In addition, the object F1 scores between dev-test and test-std sets were similar, implying that performance in dev-test can be used to predict performance on the test-std set.


Due to time limitations, we did not incorporate the scene graph features and the indicator feature of whether an object is mentioned in the previous system turn (prev\_mention). Table~\ref{tab:eval} shows that these features have brought us further improvement post evaluation. In particular, solely adding the prev\_mention feature boosted the F1 from 0.674 to 0.728. Intuitively, if an object mentioned in the previous system turns, the same object has a higher chance of being discussed in subsequent dialog turn. On the other hand, incorporating scene graphs either using attention bias or relation-aware self-attention only yielded marginal improvement, which might be due to the limited relational information in the provided scene graphs. Using more relational information deserves further investigation in the future.

\subsection{Error Analysis}
We identify two salient types of error. A frequent type of error is exemplified in Table \ref{tab:error1} where the model fails to understand the positional relationship between objects and background objects (e.g. walls, shelves, stands, racks, etc.) in the scene. Instead, the model classifies similar objects in other positions, as mentioned. The provided scene graphs are less helpful in such situations due to the lack of background objects. Another type of error relates to the complex and cluttered nature of some scenes. As demonstrated in Table~\ref{tab:error2}, some mentioned objects are very far away from the observer or have bounding boxes overlapping with other objects, making the processing of visual features difficult and inaccurate.

\begin{table}
\begin{center}
\begin{tabular}{  p{25mm} | p{35mm}  } 
 \multicolumn{2}{c}{\includegraphics[width=0.6\columnwidth]{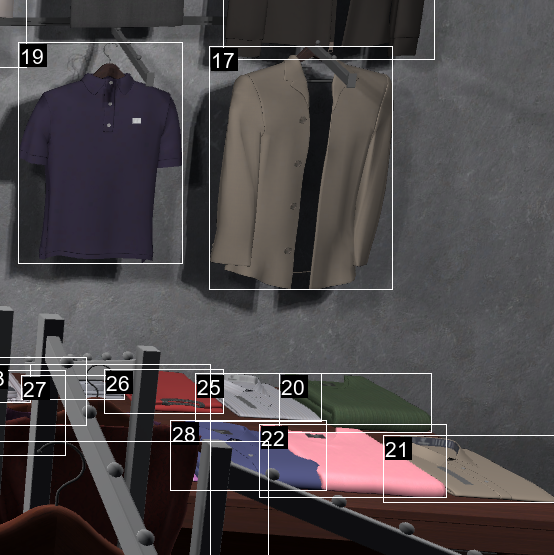}} \\
 \hline
 User utterance & The purple t-shirt hanging on the wall.\\
 \hline
 Predicted object mentions & 28\\
 \hline
 Groundtruth object mentions & 19\\
 \hline
\end{tabular}
\caption{\label{tab:error1} An example of an error where the model cannot identify a mentioned object on the wall. The image is cropped to include only the relevant region.}
\end{center}
\end{table}

\begin{table}
\begin{center}
\begin{tabular}{  p{25mm} | p{35mm}  } 
 \multicolumn{2}{c}{\includegraphics[width=0.6\columnwidth]{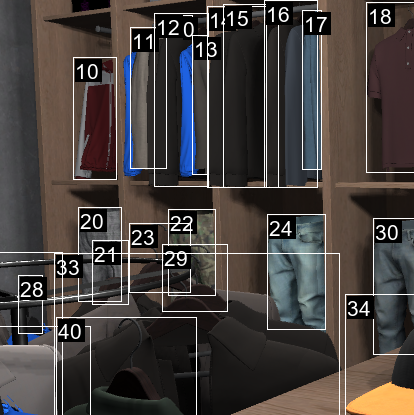}} \\
 \hline
 User utterance & Can I get the specs on the red and white sportsman jacket and that black one on the rack?\\
 \hline
 Predicted object mentions & 4, 10\\
 \hline
 Groundtruth object mentions & 4, 29\\
 \hline
\end{tabular}
\caption{\label{tab:error2} An example of an error where the model cannot identify a mentioned object on a cluttered rack. Note that the object 29 is barely visible. The image is cropped to include only the relevant region.}
\end{center}
\end{table}

\section{Conclusion}
\label{sec:conclude}
We propose a UNITER-based model addressing multimodal coreference resolution. Our model incorporates multimodal inputs, including dialog history, raw images, KB entries, scene graphs, and indicator features. Experiments show that our approach significantly outperforms the GPT2 baseline and achieved second place on multimodal coreference resolution in the DSTC10 challenge. 

A limitation of our approach is that we do not further train the pre-trained visual encoder and textual KB entry encoder. Finetuning them with the ground truth KB entries might yield better performance. Also, further efforts can be made to model the relationship between objects and the background, such as extracting and incorporating more ROI features. We leave such explorations as future work.
\bibliography{cite}

\end{document}